\documentclass[conference]{IEEEtran}
\makeatletter
\def\ps@headings{%
\def\@oddhead{\mbox{}\scriptsize\rightmark \hfil \thepage}%
\def\@evenhead{\scriptsize\thepage \hfil \leftmark\mbox{}}%
\def\@oddfoot{}%
\def\@evenfoot{}}
\makeatother
\IEEEoverridecommandlockouts
\usepackage{cite}
\usepackage{amsmath,amssymb,amsfonts}
\usepackage{algorithmic}
\usepackage[ruled,vlined]{algorithm2e}
\usepackage{graphicx}
\usepackage{textcomp}
\usepackage{xcolor}
\usepackage{comment}
\usepackage{multirow,multicol}

\def\BibTeX{{\rm B\kern-.05em{\sc i\kern-.025em b}\kern-.08em T\kern-.1667em\lower.7ex\hbox{E}\kern-.125emX}}

\def\BibTeX{{\rm B\kern-.05em{\sc i\kern-.025em b}\kern-.08em
    T\kern-.1667em\lower.7ex\hbox{E}\kern-.125emX}}
\begin{document}

\title{An Adversarial Attack Defending System for Securing In-Vehicle Networks \\
%\thanks{This paper is under review. Please do not distribute it.}
}

\author{\IEEEauthorblockN{Yi Li}
\IEEEauthorblockA{
\textit{University of South Florida}\\
Tampa, FL, USA \\
yli13@mail.usf.edu} 
\and
\IEEEauthorblockN{Jing Lin}
\IEEEauthorblockA{
\textit{University of South Florida}\\
Tampa, FL, USA\\
jinglin@mail.usf.edu}
\and
\IEEEauthorblockN{Kaiqi Xiong}
\IEEEauthorblockA{
\textit{University of South Florida}\\
Tampa, FL, USA \\
xiongk@usf.edu}
}

\maketitle

\begin{abstract}
In a modern vehicle, there are over seventy Electronics Control Units (ECUs). For an in-vehicle network, ECUs communicate with each other by following a standard communication protocol, such as Controller Area Network (CAN). However, an attacker can easily access the in-vehicle network to compromise ECUs through a WLAN or Bluetooth. Though there are various deep learning (DL) methods suggested for securing in-vehicle networks, recent studies on adversarial examples have shown that attackers can easily fool DL models. In this research, we further explore adversarial examples in an in-vehicle network. We first discover and implement two adversarial attack models that are harmful to a Long Short Term Memory (LSTM)-based detection model used in the in-vehicle network. Then, we propose an Adversarial Attack Defending System (AADS) for securing an in-vehicle network. Specifically, we focus on brake-related ECUs in an in-vehicle network. Our experimental results demonstrate that adversaries can easily attack the LSTM-based detection model with a success rate of over 98\%, and the proposed AADS achieves over 99\% accuracy for detecting adversarial attacks.

\end{abstract}


\section{Introduction}\label{sec:introduction}Nowadays, a modern vehicle, such as a connected and autonomous vehicle (CAV), contains a great number of Electronics Control Units (ECUs) that communicate with each other based on a standard communication protocol, such as Controller Area Network (CAN), in a vehicle network. With the development of CAVs, more and more ECUs are added in a vehicle for advancing driving functionalities, such as lane changing, parking assistance, and emergency braking~\cite{mudalige2015efficient}. All these features may require a vehicle to communicate with itself, other vehicles, and infrastructures, such as Vehicle-to-Vehicle (V2V) and Vehicle-to-Infrastructure (V2I) communication~\cite{wang2019survey}. 

Securing in-vehicle networks is as important as ensuring the security of Vehicle-to-Everything (V2X) communication. Researchers have shown that ECUs in an in-vehicle network are vulnerable to remote attacks~\cite{huang2018vehicle, avatefipour2018state}. An attacker can easily access an in-vehicle network to compromise ECUs through a WLAN or Bluetooth since ECUs are all connected and vice versa. Thus, in-vehicle networks are vulnerable to a variety of attacks, such as intrusion attacks and false data injection attacks (FDIA)~\cite{carsten2015vehicle}.
Deep learning methods have been widely used to detect attacks in in-vehicle networks~\cite{khan2019vehicle, kang2016intrusion}. However, recent studies have shown that deep learning models are usually vulnerable to adversarial examples~\cite{szegedy2013intriguing}. Adversaries can manipulate input data and force a trained model to produce mis-classified outputs. This makes a deep learning-based detection model useless. Hence, it is crucial to building a robust deep learning model against adversarial examples.

Current studies are mostly focused on developing and defending adversarial attacks in image datasets, such as crafting adversarial examples based on the MNIST dataset and attacking traffic signs in a CAV system~\cite{chakraborty2018adversarial}. 
In this paper, we explore two adversarial attacks: Fast Gradient Sign Method (FGSM) attack and Basic Iterative Method (BIM) attack on sequential datasets. To show the limitation of the machine learning based detection model, we first adopt the method used in~\cite{khan2019vehicle} to build an LSTM-based detection model that can detect false data injection attacks with a success rate of over 98\%. Then, we show that FGSM and BIM attacks can easily bypass this high performance LSTM-based detection model among 99\% of the time. To countermeasure against these adversarial attacks, we propose an Adversarial Attack Defending System (AADS) for in-vehicle networks, where AADS can build a robust LSTM-based detection model that able to detect both false data injection attack and two adversarial attacks efficiently. 

Our main contributions are summarized as follows:
\begin{enumerate}
\item To demonstrate the limitation of the LSTM-based detection model, we implement two adversarial attacks for attacking the LSTM-based detection model. Our experimental results demonstrate that adversaries can easily attack the LSTM-based detection model with a success rate of over 98\%.
\item To overcome these adversarial attacks, we propose AADS to detect adversarial attacks efficiently. To be precise, we countermeasure two adversarial attacks, FGSM and BIM, by adversarially retrianing our detection model iteratively such that we obtain a robust
detection model to defend aganist both attacks.
\end{enumerate}

The rest of the paper is organized as follows. Section~\ref{sec:related} \textcolor{black}{discusses related work about in-vehicle network security and adversarial attacks. Section~\ref{sec:attack} presents adversarial attack models. In Section~\ref{sec:methodology}, we give the design of the proposed AADS. Then, in Section~\ref{sec:evaluation}, we discuss the evaluation of AADS, such as detection accuracy, adversarial attack success rate, and results of defending the adversarial attacks.} Last, the conclusion of our studies and future work are presented in Section~\ref{sec:conclusion}.

\section{Related Work}\label{sec:related}An in-vehicle network is vulnerable to many different types of attacks due to several intrinsic vulnerabilities of a CAN bus, such as broadcast transmission, no authentication, and no encryption. The attackers can access in-vehicle networks through interfaces like Bluetooth and Wi-Fi to perform attacks, such as FDIA, replay attack, and Denial of Service (DoS) attack~\cite{liu2017vehicle}.

Recently, machine learning methods have been applied in intrusion detection for in-vehicle networks. Kang et al.~\cite{kang2016intrusion} proposed an intrusion detection system (IDS) using a deep neural network (DNN) to secure in-vehicle networks. A pre-trained unsupervised deep belief network was used to initialize the parameters of their DNN model in order to achieve better accuracy. Kuwahara et al.~\cite{kuwahara2018supervised} focused on the observation of CAN messages in a fixed time window to detected intrusions. They adopted both supervised and unsupervised learning to perform intrusion detection.

Khan et al~\cite{khan2019vehicle} proposed an in-vehicle false information attack defending framework using machine learning and adopted SDN in their framework. Their highest classification accuracy is 95\% with a precision and recall of 0.95 and 0.87, respectively.

\textcolor{black}{Deep learning methods have been found to be extremely successful for a variety of fields. However, researchers have recently discovered that the deep learning models can be compromised by adversarial attacks~\cite{szegedy2013intriguing}. Chakraborty et al.~\cite{chakraborty2018adversarial} surveyed different types of adversarial attacks and present some existing countermeasures against adversarial attacks. They discussed the efficiency and challenges of those countermeasures. Most existing studies on adversarial attacks focus on attacking images~\cite{carlini2017towards, kurakin2016adversarial, goodfellow2014explaining}.}

More recently, there are some studies regarding adversarial attacks in the CAV domain. They mainly focus on image aspects as well, such as vehicle detection and object detection~\cite{qayyum2019securing}. Cao et al.~\cite{cao2019adversarial} proposed an adversarial sensor attack on LiDAR sensors to fool the LiDAR-based perception into detecting wrong objects and making wrong decisions in an autonomous vehicle. Their new attacking method increased the success rate by 2.65 times on average. In this paper, we focus on attacking an in-vehicle network. We study the sequential dataset instead of image dataset. We consider two adversarial attacks, FGSM and BIM, as our threat models. To the best of our knowledge, our study is the first to apply adversarial attacks on in-vehicle networks.

\section{Attack Model}\label{sec:attack}Deep learning methods have been widely applied to a variety of domains, including image classification, Natural Language Processing (NLP), speech recognition, CAVs, and malware detection. Ensuring the security of deep learning methods used in those domains has become a crucial task, especially in the area of security-critical environments, such as CAVs.
Szegedy et al.~\cite{szegedy2013intriguing} first found that the neural network models can be compromised by adversarial attacks. 
That is, attackers can create adversarial examples to limit the use of neural networks.
Similarly, adversarial examples can be used by attackers to cause a vehicle to take unexpected actions and result in disastrous consequences. \textcolor{black}{In this section, we first briefly present the FDIA model adopted from~\cite{khan2019vehicle} to attack ECUs. Then, we introduce two adversarial attack methods to attack the LSTM-based detection model.}

\subsection{False Data Injection Attack (FDIA)}
An adversarial attacker has various ways to modify the original data. One easy way to modify the original data is to add some uniform random noise to it. We first observe the normal range for each signal. Then, we generate a random value that follows the uniform distribution within the interval of the normal range. The equation of the FDIA model is shown as follows:
\begin{equation}
X_{i}^{'} = X_i + \delta_i, \,
\end{equation}
where $\delta_i \sim \mathcal{U}(-X_i,\, X^{max}_i - X_i)$ is a uniform random value added to the normal data to create the adversarial data, $X^{max}_i$ is the maximum value of a signal, $i$ denotes the signal number, and $X_i^{'}$ is the adversarial data of $i$th signal modified by the attacker. For example, $X^{'}_6$ is the adversarial data of the 6th signal modified by the attacker. Detailed description of the dataset can be found in~\cite{khan2019vehicle}.
%Detailed description of the dataset can be found in Section~\ref{sec:methodology}.

\subsection{Fast Gradient Sign Method (FGSM)}
\textcolor{black}{In this paper, we assume that an attacker has complete knowledge of a neural network model. That is, the attacker can access the architecture of the neural network model and all parameters from the model. This is considered as a white-box attack.}

First, let's give out some notations. Let $F(X)$ be the neural network model that takes the input $X \in \mathbb{R}^n$ for some integer $n$ and yields the output $y$. Given a valid input $X$, attackers will try to create a similar input $X^{'}$ so that the output of adversarial input does not match its original one, that is $F(X^{'}) \neq F(X)$, but $||X-X^{'}||<\eta$ for some specified small number $\eta \in \mathbb{R}$. Note that the $||*||$ is some application specific metrics for measuring the similarity between $X$ and $X^{'}$. \textcolor{black}{In this paper, the distance metric we use is L2 metric, which is a standard Euclidean distance between $X$ and $X^{'}$.}

There are two types of adversarial attacks: targeted and untargeted adversarial attacks. Targeted adversarial attacks search for an adversarial input $X^{'}$ such that it will be classified as the specified class $t$. That is, $F(X^{'}) = t, t \neq y$. On the contrary, untargeted adversarial attacks only search for an adversarial input $X^{'}$ so that $F(X^{'}) \neq y$. 

An attacker can attack the deep learning based detection model by using the fast gradient sign method~\cite{goodfellow2014explaining}. The deep learning model is obtained through gradient descent optimization to minimize the cost function $ J(X,y_{label})$, i.e., reduce the error between the true label $y_{label}$ and the predicted labels $y_{pred}$. On the contrary, an attacker wants to increase the error, and a simple but effective way to do it is increasing the error by gradient ascent of the cost function $\nabla _{X}J(X,y_{label})$. That is, an attacker can simple add noise in the direction of gradient ascent of the cost function $ sign(\nabla _{X}J(X,y_{label})).$ Hence, $X^{'}$ can be calculated using FGSM as follows:
\begin{equation}
X^{'} = X + \epsilon \times sign(\nabla _{X}J(X,y_{label})),
\end{equation}  
where $\epsilon$ is included to control the magnitude of the noise. In order to keep the adversarial examples $X'$ as similar as $X$, we usually try to choose a value of $\epsilon$ that is as small as possible.

\subsection{Basic Iterative Method (BIM)}
BIM is an iterative version of FGSM proposed by Kurakin~\cite{kurakin2016adversarial}. % since FGSM is designed to be fast, but is not necessarily the optimal solution to create adversarial examples. 
BIM generates the adversarial examples iteratively using a smaller step $\alpha$. First, we set up the initial:
\begin{equation}
X^{'}_{0} = X,
\end{equation} 
and then $X^{'}_{i+1} $ is calculated as follows:
\begin{equation}
X^{'}_{i+1} = Clip_{X, \epsilon}\{ X^{'}_{i} + \alpha \times sign(\nabla _{X}J(X,y_{label}))\}, \\
\end{equation}  
where $\alpha$ is the step size and $Clip_{X, \epsilon}\{*\}$ is the element-wise clipping of X to make the clipped value belong to the range of $[X_{i,j} - \epsilon, X_{i,j} + \epsilon]$. This ensures that the generated adversarial examples are in the original range of the input $X$ so that it cannot be easily detected by a simple outlier detector.

\begin{figure}[t]
	\centering
	\includegraphics[width=\columnwidth]{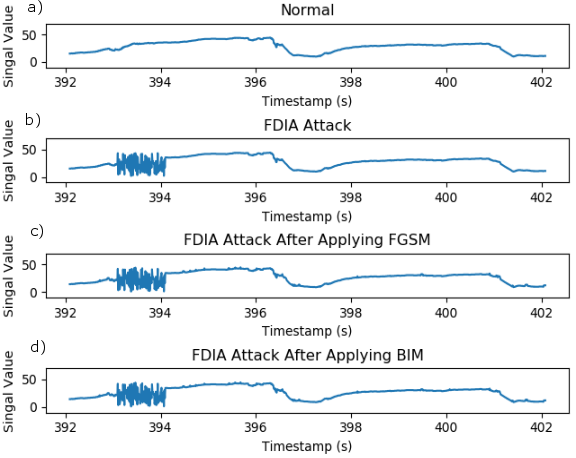}  %\textwidth
	\caption{Attack on TQI\_ACOR signal. (a) Normal signal that is correctly classified as "Normal" by the LSTM-based detection model; (b) FDIA attack signal that is correctly classified as "Attack" by the LSTM-based detection model; (c) FDIA attack signal being misclassified as "Normal" after applying FGSM to it even though signal in (c) is similar to signal in (b). (d) FDIA attack signal being misclassified as "Normal" after applying BIM to it.} 
	\label{fig:atkExample}
\end{figure}

To craft adversarial data using FDIA, we observe the normal range for each signal from the DBC file for KIA Soul and generate a random value that follows the uniform distribution within the interval of the normal range. For the sample that has a period of 10 seconds, we then randomly choose 1 second to attack. Figure~\ref{fig:atkExample}(b) shows an example of FDIA attack to the normal signal in Figure~\ref{fig:atkExample}(a). A LSTM-based detection model proposed in~\cite{khan2019vehicle} can easily detect this attack model with nearly 100\% success rate. However, such a type of deep learning based detection model itself is vulnerable to adversarial attacks. For instance, we can modify the attack signal in Figure~\ref{fig:atkExample}(b) slightly with fast gradient sign method described in subsection B. As shown in Figure~\ref{fig:atkExample}(c), there is still attack at 393-394 seconds and its signal is quite similar to one shown in Figure~\ref{fig:atkExample}(b). Intuitively, we can assume if LSTM-based detection model can correctly identify the attack signal in (b), it likely ables to detect the attack signal in (c). However, the LSTM-based detection model unable to detect it; i.e., it classifies signal in (b) as "Attack" but misclassifies signal in (c). as "Normal." Similarly, in the case of BIM, Figure~\ref{fig:atkExample}(d), the LSTM-based detection model fails to recognize that signal in (d) is the attack signals.

\section{Methodology}\label{sec:methodology}CAN is a standard bus communication protocol that has been widely used for in-vehicle network communication. It uses a broadcasting method to transmit messages from one ECU to other ECUs on the bus. The source and destination information are not included in the transmitted messages. Thus, it is vulnerable to attacks due to a lack of authentication. An attacker can easily modify the data on the CAN bus, and this can result in vehicle malfunctions. In this section, we first describe the dataset and data processing method used in this study. Then, we introduce our AADS in details.

\subsection{Data Processing}
In order to examine the efficacy of proposed AADS, the dataset from real in-vehicle CAN bus is used in this paper. It is collected from the KIA SOUL by Hacking and Countermeasure Research Lab~\cite{lee2017otids}. 
In order to better understand and process the dataset, the raw CAN data frames need to be decoded. We use a generic KIA DBC file decoder from OpenDBC repository to decode the dataset~\cite{opendbc}. OpenDBC is created by CommaAI with a repository of DBC files for different models of vehicles. The DBC file is usually used to describe the data over a CAN bus. It has the information to decode CAN data frames. For instance, we can use the scale and offset values in the DBC file to convert raw bits of data into signal values. In the raw dataset, the CAN IDs (CID) are in hexadecimal format whereas the Message IDs (MID) are decimal in the DBC file. Therefore, we first need to convert the CIDs in the raw dataset to match the MIDs in the DBC file for decoding. 

Each CAN ID has a specific ECU associated with it, and each ECU has multiple signals. Each signal has its own decoding information like scale and offset. For the comparison purpose, we explore 20 signals associated with five ECUs the same as Khan et al.\cite{khan2019vehicle}. These ECUs are Motor Driven Power Steering (MDPS), Engine Management System (EMS), Anti-lock Braking System (ABS), Electronic Parking Brake System (EPB), and Electronic Stability Control System (ESC). MDPS and EMS ECUs send CAN messages/frames to ABS, EPB, and ESE. EMS broadcasts four types of CAN messages (MSG), including EMS11, EMS12, EMS14, and EMS16, whereas MDPS ECU broadcasts SAS11 CAN messages. \textcolor{black}{Each type of CAN messages contains different signals, such as TQI\_ACOR, TQFR, N, TQI, TPS, VB, etc.} 
The detailed information of message type and signals as well as the detailed information for decoding the raw CAN data frames can be found in~\cite{khan2019vehicle}.

Since we only investigate brake related ECUs, we first need to filter out all unrelated data. After filtering, we have 952,101 rows of CAN data frames. The number of CAN data frames for each CAN ID is evenly distributed. We use 20 signals presented in~\cite{khan2019vehicle}.

Then, the dataset is divided based on the timestamps. We choose each 10 seconds as one sample. For example, the first sample lasts from the 1st second to the 10th second, and the second sample lasts from the 2nd second to the 11th second, and so on. Thus, we creates 1894 samples in total. Among these samples, half of them are randomly selected to craft adversarial data.

\subsection{AADS Architecture}
In this study, we propose an adversarial attack defending system to defend against adversarial attacks. We create a malicious ECU that is connected to the CAN bus.  
There are six ECUs connected to the CAN bus. EMS and MDPS ECUs can only broadcast CAN messages, whereas ABS, ESC, and EPB ECUs can only receive CAN messages. The malicious ECU can both broadcast and receive CAN messages.

\subsubsection{LSTM-based Detection Model}
In the AADS, we build an LSTM-based detection model to detect FDIA by following~\cite{khan2019vehicle}. 
We use 20 signals as input features, and each input feature is a sequence of decoded CAN signal values. In our model, the LSTM layer contains 128 neurons. 
Then, an output layer with only one neuron follows the LSTM layer. This is because we are dealing with a binary classification problem, whose output is either 1 (Attack) or 0 (Normal). The activation function we use for this layer is sigmoid. The cost function we use is binary cross-entropy. The optimization function we choose is Adam~\cite{kingma2014adam}, which is an adaptive learning rate optimization algorithm. It computes individual learning rates for different parameters. 

In our dataset, we have 1894 instances in total. Among these instances, we use 80\% of the dataset for training, 10\% for validation, and 10\% for testing. To be precise, we have 1516 samples for training, 189 samples for validation, and 189 samples for testing. The total trainable parameters are 76,417.

\subsubsection{Adversarial Attacks Defending Scheme}
Adversarial attacks can compromise the trained machine learning model to produce incorrect classification results. In a scenario like CAVs, it can cause a deadly outcome if the system failed to perform correctly. In this paper, we explore two adversarial attacks, FGSM and BIM. First, we generate adversarial examples using both FGSM and BIM on the test dataset to see the efficiency of compromising the trained LSTM model. Then, we propose an adversarial retraining method to make the LSTM model more robust against both the FGSM and BIM attacks.

In order to build a robust LSTM detection model, we iteratively re-train the LSTM model using both the adversarial examples and the original training samples.  
Let $S$ denote original samples, $S^{'}$ denote samples after adversarial attacks on $S$, let $N$ denote the batch size, and let $i$ be the $i$-\textit{th} iteration of re-training. Algorithm~\ref{alg:alg1} shows the method of our adversarial attack defending scheme. For iteration $i$, the detailed steps of iterative re-training are 
1) randomly choose $N$ instances $S_{i}$ from the training dataset; 
2) generate $N$ adversarial examples $S^{'}_{i}$ using the adversarial model; 
3) add $N$ adversarial examples $S^{'}_{i}$ to adversarial repository $S^{repo}$;
4) randomly choose N adversarial samples $S^{adv}_{i}$ from the repository $S^{repo}$;
5) combine the selected $N$ adversarial examples $S^{adv}_{i}$ and the $N$ original instances $S_{i}$, so we have $2N$ samples for training. After re-training, we validate on validation dataset $V$ and check whether the stopping criteria is met. If it is met, we stop the re-training process. Otherwise, the process repeats. Each iteration produces a new LSTM model $M_i$. When process stops we save the final LSTM model$M_f$, and we evaluate $M_f$ using the test dataset to determine the robustness of $M_f$.

\begin{algorithm}
\SetAlgoLined
\KwResult{Robust LSTM Model $M_{f}$}
 \While{Training dataset is not empty}{
  Randomly choose $N$ instances $S_{i}$ from the training dataset\;
  Use adversarial model to generate $S_{i}^{'}$\;
  Add $S_{i}^{'}$ to adversarial repository $S^{repo}$\;
  Randomly choose $N$ adversarial samples $S_i^{adv}$\;
  Train on $S_{i} + S_{i}^{adv}$, Validate on $V$ }
 \label{alg:alg1}
 \caption{Adversarial Defending Algorithms}
\end{algorithm}

\section{Experimental Evaluation}\label{sec:evaluation}

%To conduct the experiment, we first set up the SDN-based in-vehicle network on GENI~\cite{GENI-Berman}, which is a nationwide suite of infrastructure supporting ``at scale" research in networking, distributed systems, security, and novel applications. We use Floodlight as our SDN controller~\cite{floodlight}. The SDN-based in-vehicle network consists of six ECUs, including EMS, MDPS, ESC, EPB, ABS, and a malicious ECU. EMS, MDPS, and the malicious ECU will broadcast CAN messages. Only EMS and MDPS broadcast authentic messages and malicious ECU broadcasts adversarial messages. ESC, EPB, and ABS will only receive messages.
In this section, we first evaluate the LSTM-based detection model. Then, we evaluate the effectiveness of FGSM and BIM attack followed by the evaluation of our adversarial attack defending scheme. %Last, we evaluate our mitigation scheme in SDN network.

\subsection{LSTM-based Detection Model Evaluation}
The LSTM-based detection model has an input layer, a LSTM layer with 128 neurons, and an output layer. The number of epochs is set to 50. The training process is carried out on GAIVI2, which is a computer cluster with large scale parallel computing capabilities. %\textcolor{blue}{ Each node has the hardware: GeForce GTX TITAN X major: 5 minor: 2 memoryClockRate(GHz): 1.076.}
GAIVI2 has a TensorFlow version of 2.0. We test with different optimizers, such as SGD, Adam, RMSprop, and Adagrad.  Table~\ref{tab:eval} shows the evaluation of accuracy, recall, precision, F1 score, and training time for different optimizers. The accuracy is the accuracy of the unseen test set, and the Time column reports the time used to train the model in seconds.

\begin{figure}[t]
	\centering
	\includegraphics[width=3.3in]{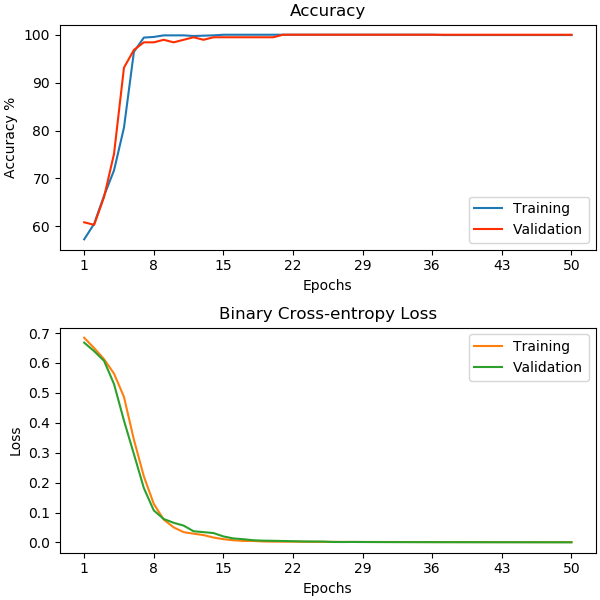}  %\textwidth
	\caption{Evaluation of accuracy and binary cross-entropy loss}
	\label{fig:acc_loss}
\end{figure}

\begin{table}[h]
    \caption{Evaluation of Different Batch Sizes}
    \small\addtolength{\tabcolsep}{-1.8pt}
    \begin{tabular}{| c | c | c | c | c | c |} 
    %\toprule %inserts double horizontal lines
    \hline
     {\textbf{Optimizer}} &{\textbf{Accuracy}} & {\textbf{Recall}}  & {\textbf{Precision}} & {\textbf{F1 Score}} & {\textbf {Time(s)}} \\ \hline
      RMSprop & 98.42\% & 0.98 & 0.98 & 0.98 & 83.96\\ \hline
      Adam    & 99.47\% & 0.99 & 0.99 & 0.99 & 82.66\\ \hline
      Adagrad & 73.16\% & 0.73 & 0.80 & 0.72 & 83.32\\ \hline
      SGD     & 64.74\% & 0.65 & 0.68 & 0.63 & 83.38\\ \hline
      %\bottomrule
    \end{tabular}
    \label{tab:eval}
\end{table}

\begin{figure*}[t]
	\centering
	\includegraphics[width=6.2in]{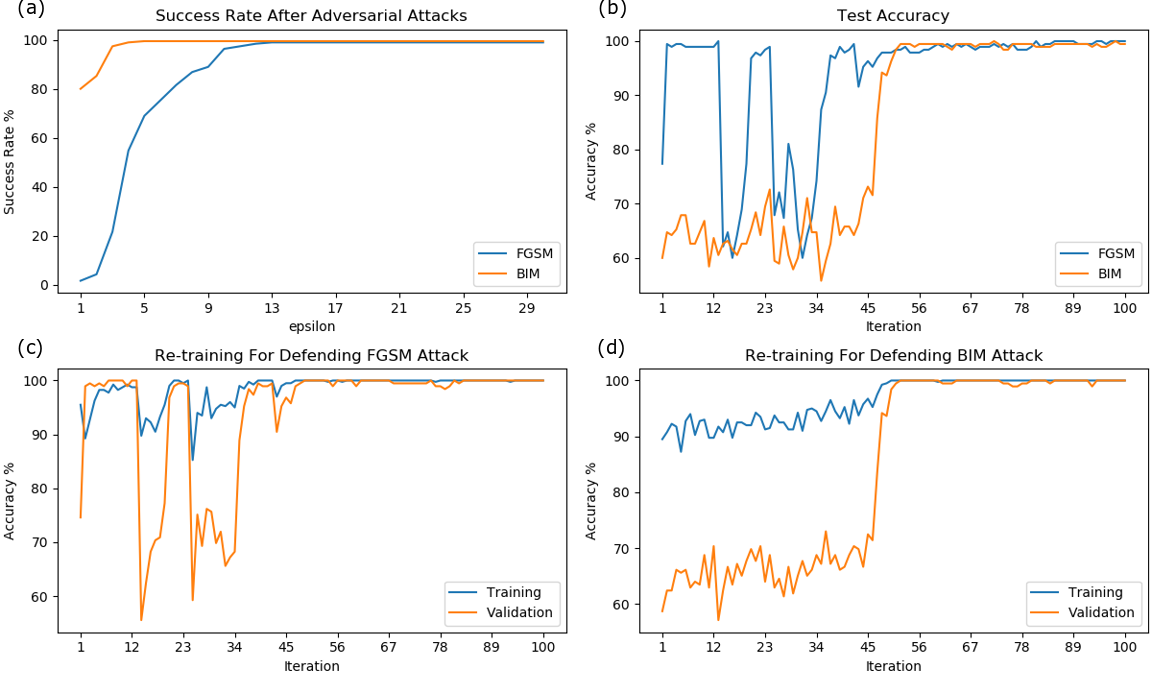}  %\textwidth
	\caption{Evaluation of adversarial attack defending scheme}
	\label{fig:res}
\end{figure*}

We can see from Table~\ref{tab:eval} that with Adam optimizer, we have achieved the best performance with an accuracy of 99.47\%. The recall, precision, and F1 score for the Adam optimizer are all 0.99. RMSprop optimizer has the second-best performance with an accuracy of 98.42\%, and both recall and precision are 0.98. The time for training the model by using Adam, RMSprop, Adagrad, and SGD optimizer are 82.66 seconds, 83.96 seconds, 83.32 seconds, and 83.38 seconds, respectively. We can see that the training time is very short. This makes AADS possible to update the model every two minutes and perform the detection in real-time. \textcolor{black}{Compare to Khan et al.~\cite{khan2019vehicle} our approach has a better performance. Khan et al. have the highest accuracy of 95\%, precision of 0.95, and recall of 0.87, while our approach has the highest accuracy of 99.47\%, precision of 0.99, and recall of 0.99.}

Figure~\ref{fig:acc_loss} shows the final results of the LSTM-based detection model. The training process runs for 50 epochs. On the top of Figure~\ref{fig:acc_loss}, it can be seen that we achieved a training accuracy of 99.93\% and the validation accuracy of 99.98\%. On the bottom of Figure~\ref{fig:acc_loss}, we report the binary cross-entropy loss, where the losses are 0.0012 and 0.00096 for training and validation respectively.

\subsection{Adversarial Defending Scheme Evaluation}
In this subsection, we demonstrate the limitation of the LSTM-based detection model by implementing two adversarial attacks for attacking the LSTM-based detection model obtained in the previous section. Then, we introduce our defending method. 

We generate the adversarial example and use it to show the limitation of LSTM-based detection model. In order to keep the adversarial examples as similar as the original input, we try to choose a $\epsilon$ as small as possible. Figure~\ref{fig:res}(a) shows the attack success rates on the test dataset using both FGSM and BIM with different $\epsilon$ values. In the BIM attack, we find that when we set $\alpha = 1$ and iterate five times, we can achieve the best results. We can see from Figure~\ref{fig:res}(a) that when $\epsilon = 5$, BIM has reached the best success rate of 99.47\%. Similarly, when $\epsilon = 13$, FGSM has achieved the best success rate of 98.42\%. In general, the BIM attack is more efficient than the FGSM attack to compromise our LSTM model. Table~\ref{tab:aa} shows the success rates, accuracies, best $\epsilon$, and time used to craft adversarial examples on the test dataset for both FGSM and BIM.
\begin{table}[h]
    \centering
    \caption{Evaluation After Adversarial Attacks}
    %\renewcommand{\arraystretch}{1.3}
    %\small\addtolength{\tabcolsep}{-1.0pt}
    \begin{tabular}{| c | c | c | c | c |} 
    %\toprule %inserts double horizontal lines
    \hline
     {\textbf{Method}} & {\textbf{Success Rate\%}} &{\textbf{Accuracy\%}}  & {\textbf{$\epsilon$}} & {\textbf{Time (s)}}\\ \hline
     FGSM& 98.42\% & 1.58\% & 12 & 9.24\\ \hline
     BIM  & 99.47\%  & 0.53\% & 5 & 37.35\\ \hline
      %\bottomrule
    \end{tabular}
    \label{tab:aa}
\end{table}

In our adversarial attack defending scheme, we apply iterative re-training to countermeasure adversarial attacks. We choose batch size $N = 200$ and iterate 100 times. Figure~\ref{fig:res}(c) shows the re-training accuracies against the FGSM attack at each iteration. The accuracies are evaluated on the validation set. We can see that when we run approximately 62 iterations, the training and validation accuracies become stable. This indicates that the FGSM attack can no longer affect our re-trained LSTM model. In the early iterations, there are some large fluctuations, which is because for each iteration we randomly choose 200 samples from the training dataset and it takes some iterations to cover all the training dataset to be selected. Figure~\ref{fig:res}(d) shows the re-training accuracies against the BIM attack. When we run around 56 iterations, the training and validation accuracies become stable. Figure~\ref{fig:res}(b) compares the test accuracy of iterative re-training against FGSM and BIM, respectively. It shows that BIM attack performs better than FGSM to fool the LSTM model for the first 56 iterations. However, after more iterations, our LSTM model becomes more robust. Neither attack can reduce the performance of the retrained LSTM. Our results demonstrate that our adversarial attack defending method can successfully defend against both FGSM  and BIM by iterative re-training.

\section{Conclusion}\label{sec:conclusion}In this research, we proposed an adversarial attack defending system for in-vehicle networks. To demonstrate the need for such a system for any deep learning-based detection model and to show the limitation of the general deep learning-based detection model, we follow a recent paper by Khan et al.~\cite{khan2019vehicle} to build an LSTM-based detection model. This LSTM-based detection model can successfully detect an attack signal generated by FDIA over 98\% time. This may seem to be a good detection model initially; however, we are able to attack this LSTM-based detection model using FGSM and BIM attacks, respectively. The accuracies of the LSTM-based detection model are only 1.58\% and 0.53\% under FGSM and BIM attacks, respectively. This basically makes the LSTM-based detection model useless. This clearly indicates the drawback of using a deep learning-based detection model. Then, we present our proposed defending scheme against these attacks. The experimental results demonstrated that after our adversarial attack defending scheme is used, the re-trained LSTM-based detection model becomes robust and is no longer affected by both FGSM and BIM attacks. Actually,  we achieved an accuracy of 99.47\% for detecting FDIA. In the future, we will explore more complex adversarial attack models to craft adversarial CAN messages, as well as to formalize the security guarantee.

\bibliographystyle{IEEEtran}
\bibliography{reference}

\end{document}